\documentclass[10pt,twocolumn]{article}

\usepackage{times}
\usepackage{graphicx}
\usepackage{amsmath}
\usepackage{tikz}
\usetikzlibrary{positioning, arrows.meta}
\usepackage{multirow}
\usepackage{booktabs}

\usepackage{xcolor}
\usepackage{apacite}
\usepackage[english]{babel}
\usepackage{setspace}

\newif\ifshowchanges
\showchangesfalse   

\ifshowchanges
  \newcommand{\added}[1]{\textcolor{blue}{#1}}
  \newcommand{\removed}[1]{}
\else
  \newcommand{\added}[1]{#1}
  \newcommand{\removed}[1]{}
\fi

\begin{document}

\csname title\endcsname{Integrating Survival-Based Aging Models with Data-Driven RUL Prognostics}
\author{%
    Abhishek Srinivasan$^{1}$, Juan Carlos Andresen$^{1}$,
    Sepideh Pashami$^{2,4}$, and Anders Holst$^{3,4}$\\[0.5em]
    \small
    $^{1}$Traton AB, Södertälje, Sweden\\
    $^{2}$Halmstad University, Halmstad, Sweden\\
    $^{3}$KTH Royal Institute of Technology, Stockholm, Sweden\\
    $^{4}$RISE Research Institutes of Sweden AB, Stockholm, Sweden
}
\date{}

\maketitle

\begin{abstract}

Predictive maintenance requires reliable remaining useful life (RUL) estimation. Existing methods mainly follow two paradigms: wear-based aging models that capture cumulative degradation and sensor-driven data models that reflect instantaneous health conditions, each providing only partial information.
In this work, we propose a probabilistic fusion framework that integrates wear-based and sensor-based prognostic components through failure probability distributions. Based on explicit structural assumptions linking wear, latent health, sensor observations, and failure, we derive a principled combination rule that enables uncertainty-aware integration with adaptive weighting of the components.
Experimentally, we assess this combination rule by learning the wear-based component using a parametric survival model and the sensor-based component using a 1D convolutional neural network (1D-CNN) with a post-hoc uncertainty model. Evaluation on multiple N-CMAPSS datasets demonstrates that the fused model improves point accuracy, preserves the C-index, and produces narrower yet well-calibrated prediction intervals compared to either component alone. The results highlight the complementary roles of wear-based survival model and sensor-based deep learning model, and show that their probabilistic integration provides a structured pathway toward more robust and consistent prognostics over the life-time.
\end{abstract}

\section{Introduction}

In recent years, the reliability of engineering systems has become increasingly important, particularly as modern society depends heavily on complex and safety-critical assets such as aircraft engines, energy systems, automobiles, and industrial manufacturing equipment. Ensuring safe and efficient operation under demanding conditions has become a central concern in engineering practice. In this context, maintenance approaches have progressively shifted from reactive and preventive maintenance toward predictive maintenance, which seek to anticipate failures using condition monitoring models \cite{murtaza2024paradigm}.

Predictive maintenance relies on condition monitoring systems that infer the state of machinery during operation \removed{\cite{fink2020potential}}. These systems generally serve two distinct but related purposes: diagnostics and prognostics. Diagnostic approaches, such as anomaly detection or health index estimation, assess the current condition of the system and identify deviations from normal behavior. Prognostic approaches aim to estimate the future progression of degradation, typically in the form of a time-to-event distribution or Remaining Useful Life (RUL). While diagnostic information provides insight into the present state, effective predictive maintenance ultimately requires prognostic estimates to support maintenance scheduling, risk management, and resource planning \cite{fink2020potential}.

A wide range of approaches have been proposed for RUL estimation within the fields of reliability engineering and prognostics and health management (PHM) \removed{\cite{fink2020potential}}. Broadly, existing methods can be categorized according to their primary source of information: \added{\textit{wear-based models}, which rely on accumulated covariates, and \textit{sensor-based models}, which use instantaneous multivariate sensors}\removed{wear-based aging models and sensor-based data-driven models}\added{\cite{zheng2017long,nemani2022ensembles}}.


Wear-based models describe failure as a function of accumulated usage, age, or exposure to operating conditions. Classical reliability approaches often assume parametric lifetime distributions such as Weibull or log-normal models \cite{klein2003survival}, while semi-parametric formulations such as the Cox proportional hazards model estimate relative hazard as a function of covariates without specifying the baseline hazard \cite{kaplan1958nonparametric,cox1972regression}. Extensions of these models allow for time-varying covariates, enabling the incorporation of evolving operational factors over time \cite{cygu2023comparing}. Nevertheless, typically such time-varying covariates are engineered or aggregated to form representations of system wear rather than high-frequency sensor measurements. As a result, these models impose a coherent time-to-event structure and yield monotone survival functions, providing temporally consistent estimates of failure risk, but they often rely on simplified aging assumptions and do not fully capture the instantaneous health state reflected in real-time sensor data \cite{cao2024hybrid,cygu2023comparing}.

In contrast, sensor-based prognostic models infer degradation directly from multivariate time-series measurements collected during system operation \cite{zheng2017long}. Machine learning and deep learning methods, including convolutional neural networks (CNNs) and recurrent architectures such as LSTMs, have demonstrated strong predictive performance \cite{zheng2017long,nemani2022ensembles}. These approaches are capable of capturing complex nonlinear relationships between sensor signals and degradation patterns. However, they often treat RUL estimation as a regression problem over sensor observations and typically lack an explicit aging structure. As a result, predictions may become unstable in early stages of operation when degradation signals are weak, and temporal consistency is not guaranteed unless additional constraints are imposed \cite{nieves2022semi}.

These observations suggest that wear-based and sensor-based models capture complementary aspects of system degradation. Wear-based models provide a structured representation of long-term aging dynamics, while sensor-based models capture instantaneous variations in system condition. Nevertheless, the two paradigms have largely been developed and evaluated independently \cite{rahat2023bridging}. Many existing approaches implicitly assume that either cumulative wear variables or sensor measurements alone sufficiently explain failure behavior, limiting the ability to exploit the complementary strengths of both information sources.

The present study proposes a probabilistic fusion framework that integrates predictions from a wear-based model and a sensor-based model within a unified probabilistic formulation. The approach is motivated by an explicit structural assumption about the relationships between accumulated wear, latent health state, sensor observations, and failure occurrence. Under this assumption, predictions from both wear-based and sensor-based model can be coherently combined to obtain an improved estimate of failure probability distribution conditioned on both sources.

To evaluate the proposed approach, we compare a 1D convolutional neural network (sensor-based model), a Weibull survival model (wear-based model), and the combined framework using the N-CMAPSS dataset \cite{arias2021aircraft}. The empirical results show that the fused model achieves improved prediction accuracy, competitive c-index, and well-calibrated narrow prediction intervals compared to either model used independently. In addition, analysis of prediction behavior across the system life-time reveals that the fusion framework naturally shifts between wear-driven and sensor-driven predictions depending on the relative predictive uncertainty of each component.

\section{Related Work}

\subsection{Survival Analysis for Time-to-Failure Modeling}

Survival analysis has long been used to model time-to-failure in reliability engineering and biostatistics \cite{klein2003survival}. Classical parametric approaches assume a specific lifetime distribution, most commonly Weibull, log-normal, or exponential models, and estimate distribution parameters from failure data \cite{klein2003survival}. These models are attractive due to their interpretability and closed-form survival and hazard functions, but their validity depends strongly on distributional assumptions.

Semi-parametric and non-parametric approaches relax these assumptions. The Kaplan–Meier estimator provides a non-parametric estimate of the survival function under right-censoring \cite{kaplan1958nonparametric}, while the Cox proportional hazards (Cox-PH) model estimates relative hazards without specifying the baseline hazard function \cite{cox1972regression}. The Cox model assumes proportional hazards over time, which may not hold in systems with evolving degradation dynamics.

More recently, machine learning extensions of survival models have been proposed. Random survival forests generalize tree-based methods to improve representation power via covariates and learns via Kaplan-Meier formulation \cite{ishwaran2008random}. While deep learning formulations such as DeepSurv, uses neural Cox models to parameterize the hazard function using neural networks \cite{katzman2018deepsurv}. These approaches retain the survival modeling framework while increasing representational flexibility.

In most survival-based approaches within engineering applications, covariates typically represent accumulated usage, age, or wear-related quantities (e.g., operating hours, load exposure, cumulative stress) \cite{klein2003survival,yang2022prognostic}. Time-varying covariates may also be incorporated, but they are often aggregated or engineered features rather than high-frequency raw sensor streams \cite{cygu2023comparing}. Importantly, survival models primarily encode aging structure and long-term risk trends, rather than instantaneous condition information.

\subsection{Remaining Useful Life (RUL) Prediction}

RUL prediction has been extensively studied in prognostics and health management (PHM) \cite{fink2020potential}. Approaches can broadly be categorized into reliability-based (statistical) models, physics-based degradation models, and data-driven machine learning models.

Reliability-based approaches estimate RUL from assumed lifetime distributions or stochastic degradation processes. For example, Wiener and gamma processes have been used to model monotonic degradation trajectories and infer time-to-failure distributions \cite{deng2015degradation}. These models often provide principled uncertainty quantification but may require strong assumptions about degradation dynamics.

Physics-based models rely on first-principles understanding of degradation mechanisms, such as crack propagation (e.g., Paris’ law) or fatigue accumulation \cite{bechhoefer2008use}. While physically interpretable, such models require domain knowledge and may not scale well to complex systems with multiple interacting degradation modes.

Data-driven approaches, particularly deep learning models such as multilayer perceptrons (MLPs), convolutional neural networks (CNNs), and long short-term memory networks (LSTMs), have demonstrated strong performance on benchmark datasets such as CMAPSS and N-CMAPSS \cite{zheng2017long, nemani2022ensembles, fink2020potential}. These models typically learn mappings from multivariate sensor time-series to RUL targets. In many benchmark datasets (e.g., CMAPSS), sensor signals exhibit monotonic drifts over time, effectively acting as proxies for latent wear. However, such proxies are indirect and may not explicitly encode cumulative aging mechanisms.

Several works have attempted to incorporate stochastic degradation modeling from reliability theory with neural networks. For example, degradation processes modeled via Wiener processes with both measurable and unobservable external impacts \cite{zhang2023degradation} consider time-varying covariates and latent factors within a stochastic framework. Similarly, LSTM-based stochastic degradation models combine sequence learning with probabilistic degradation modeling \cite{zezhou2024stochastic}. These approaches integrate temporal learning with uncertainty modeling but typically focus on sensor-driven degradation rather than explicitly separating accumulated wear factors from instantaneous condition indicators.

\subsection{Bridging Survival Modeling and RUL Estimation}

Both survival analysis and remaining useful life (RUL) estimation aim to characterize how systems degrade over time, yet they approach this objective from different perspectives. Survival models emphasize time-to-event distributions and principled handling of censoring, while RUL models often focus on regression-style prediction from sensor streams.


Although survival analysis and RUL prediction address closely related time-to-failure objectives, existing hybrid approaches often combine them by balancing or averaging their contributions. Recent work such as SurvLoss \cite{rahat2024survloss} bridges RUL regression and survival modeling through a survival-consistent loss, enabling joint learning from censored and uncensored data. Other related studies have also transformed survival outputs into RUL targets or compared survival-based and regression-based models under varying censoring levels, showing the value of censoring-aware formulations for time-to-event prediction \cite{rahat2023bridging}. However, these methods still operate within a single modeling framework rather than explicitly modeling distinct data-generating factors. In contrast, our approach treats wear-based aging and sensor-based condition information as complementary but structurally separate sources, and derives a probabilistic fusion rule that combines them at the level of failure probability distributions.

\removed{Overall, relatively few works explicitly model wear-related aging factors and sensor-based condition indicators as complementary but structurally distinct sources of information within a unified probabilistic framework. 
Existing hybrid approaches often combine physics-based or stochastic degradation models with data-driven predictors, but typically do so at the level of state estimation or RUL regression rather than through an explicit factorization over wear, health, sensors, and failure \cite{cao2024hybrid,Zhang_2025jaciii,liao2016hybrid}.
Consequently, many RUL-oriented methods assume sensor data sufficiently capture degradation, while survival-based approaches rely on cumulative usage variables to explain failure risk \cite{fink2020potential,klein2003survival,zheng2017long}. This leads to a blurred distinction between long-term aging and instantaneous condition, rather than modeling them as separate but interacting components.}
\added{Existing hybrid prognostics approaches have also combined model-based or physics-informed information with data-driven RUL predictors. \citeauthor{chao2022fusing} \citeyear{chao2022fusing}, for example, used physics-based performance models to infer unobservable health-related parameters that were then combined with sensor readings as inputs to a deep neural network. \citeauthor{cao2024hybrid} \citeyear{cao2024hybrid} similarly used an exponential degradation model and extended Kalman filtering to estimate degradation-related states and parameters, which were fused with sensor-derived features in a Transformer-based RUL predictor. While these works demonstrate the value of hybrid prognostics, they primarily perform fusion at the level of input features, latent representations, state estimates, or direct RUL regression \cite{cao2024hybrid,Zhang_2025jaciii,liao2016hybrid}. In contrast, our work explicitly separates data-sources i.e., accumulated wear and sensor-based condition information as structurally distinct partial views of failure and combines the corresponding failure-time distributions through a probabilistic factorization. Thus the contribution here is not a specific hybrid model, but the principled mechanism to combine sensor based and wear based models, thereby making it possible to enhance already existing prediction models in industry. \removed{Thus, the contribution is not hybridization in general, but distribution-level, uncertainty-aware fusion grounded in explicit assumptions over wear, latent health, sensor observations, and failure.}}


\subsection{Relation to the Present Work}

In contrast to prior approaches, our work starts from an explicit assumption on data-generating process in which accumulated wear (W) influences latent health (H), which in turn affects both sensor observations (S) and failure occurrence (F). Under this structure, survival-based wear modeling and sensor-based RUL estimation correspond to partial views of the same latent process.

Rather than replacing one paradigm with the other, we derive a probabilistic fusion rule under conditional independence assumptions implied by the graphical model, enabling coherent combination of $P(F|W)$ and $P(F|S)$ into $P(F|W,S)$. The key distinction from existing stochastic degradation or hybrid deep models is that we explicitly separate wear-driven aging dynamics from sensor-driven state estimation and combine them at the level of failure probability distributions. This work makes the underlying structural assumptions and probabilistic fusion mechanism explicit.

\section{Model Assumptions}
\label{Sec: ModelAssumptions}
The assumption behind this study is that the two types of information --  accumulated wear and momentary sensor readings -- are complementary and each could therefore contribute to the prediction of when a failure will occur. The corresponding model assumption is depicted in Figure \ref{fig:model}. The wear factors ($W$) accumulated over time affect the health ($H$) of the component. The health in turn affects the momentary sensor readings ($S$). The health also affects when the component will fail ($F$).

The health ($H$) is a hidden factor, since it cannot be observed directly. \added{This gives us the freedom to select a suitable
representation of H.} We will here assume a direct and deterministic connection between this hidden health state and the failure probability, such that when the health is "used up" the component will fail. \removed{Then the probability distribution of the observable failure ($F$) can be used as a proxy for the non-observable health status ($H$) \added{(i.e. $P(F) \approx P(H)$)}.}
\added{In specific, we will let the health represent the remaining amount of equivalent operating hours (EOH) until failure. The distributions of F will also be epressed as a distribution over EOH. Then the distribution of H ("How much health is left?") will be exactly the same as the distribution of F ("How long until it fails?"). In the following dervivation we can therefore use P(F) instead of P(H).}

\removed{Using the conditional independence assumptions implied by Figure \ref{fig:model}, we derive the combined failure distribution $P(F \mid W, S)$ from the individual components. In particular, we assume that wear and sensor observations are conditionally independent given failure, i.e., $W \perp S \mid F$. Applying Bayes' rule and factorizing the joint likelihood accordingly yields the following relation:}
 

\added{This equivalence assumption allows us to express the joint likelihood in terms of observable quantities. Applying Bayes’ rule and factorizing the likelihood accordingly, we obtain:
}
\begin{figure}[t]
  \centering
\begin{tikzpicture}[
  node distance=2.2cm,
  >=Stealth,
  every node/.style={draw, circle, minimum size=6mm, inner sep=0pt, line width=0.8pt},
  every path/.style={->, line width=0.8pt}
]
  \node (W) {W};
  \node (H) [below right=1.2cm of W] {H};
  \node (F) [right=1cm of H] {F};
  \node (S) [below left=1.2cm of H] {S};

  \draw (W) -- (H);
  \draw (H) -- (F);
  \draw (H) -- (S);
\end{tikzpicture}
\caption{Model assumption of data generation process. Wear factors ($W$) influence latent health ($H$), which affects both sensor observations ($S$) and failure occurrence ($F$).}
\label{fig:model}
\end{figure}
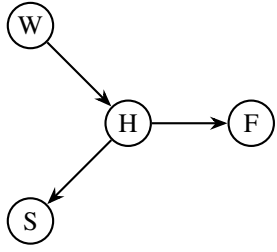
\begin{align*} \notag
    P(F \mid W, S) &\propto P(W,S \mid F) \cdot P(F)\\ \notag
    &= P(W \mid F) \cdot P(S \mid F) \cdot P(F) \\ \notag
    &\propto \frac{P(F \mid W)}{P(F)} \cdot \frac{ P(F \mid S)}{P(F)} \cdot P(F)\\ 
    &= \frac{P(F \mid W) \cdot P(F \mid S)}{P(F)}
\end{align*}


Here $P(F)$ represents the probability of failure when neither wear factors nor sensors are known. That is, lacking information on how long it has been used or how it behaves, we assume a constant hazard rate based on the MTBF (mean time between failures), resulting in an exponential failure probability distribution. We divide with P(F) to avoid factoring it twice, as it is implicitly included in both $P(F \mid W)$ and $P(F \mid S)$.

Note that we here use the Bayesian notation of using $P()$ for a probability distribution, and $\propto$ means "proportional to", which means that the expression needs to be normalized to produce a proper probability distribution in the end. All the probability distributions \removed{in this case}\added{in the final expression} are over \removed{time}\added{equivalent operating hours}, and the normalization is thus over \removed{time}\added{the same}.

\section{Method}

Based on the probabilistic derivation in the section \ref{Sec: ModelAssumptions}, the final combined equation is obtained as:
\begin{align}
\label{eq: assumption}
  P(F \mid W, S) &\propto \frac{P(F \mid W) \cdot P(F \mid S)}{P(F)}
\end{align}

the failure distribution conditioned on both wear factor ($W$) and the sensor information ($S$) consists of three components: (i) \textbf{Wear-based failure model} ($P(F \mid W)$); (ii) \textbf{Sensor-based failure model} ($P(F \mid S)$); and (iii) \textbf{Baseline failure distribution} ($P(F)$). Each component produces a failure probability density over future time.

\subsection{Wear-based failure model ($P(F \mid W)$)}
\label{method:wear}
To model failure from wear factors, we utilize Weibull survival model. Let $T$ be the time-to-failure. The Weibull distribution is defined as 
\begin{align*}
    f_W(t) = \frac{k}{\lambda} \left(\frac{t}{\lambda}\right)^{k-1}
\exp\left[-\left(\frac{t}{\lambda}\right)^k\right],
\end{align*}
where $k$ is the shape parameter and $\lambda$ the scale parameter. These parameters are estimated by optimizing to maximize the log likelihood based on the observed data. 

The distribution $P(F \mid W)$ is represented via the conditional density
over time-to-failure $\tau$ at current time $t_c$, i.e the failure density after $t_c$ given that the component has survived until $t_c$: 
\begin{align*}
f_W(\tau \mid T > t_c) = 
\frac{f_W(t_c + \tau)}{\mathcal{S}_W(t_c)},
\end{align*}
where $\mathcal{S}_w(t_c)$ is the survival probability at $t_c$. The renormalization ensures that the conditional distribution is a consistent probability distribution and that it factors for the time that has passed. 

\subsection{Sensor-based failure model ($P(F \mid S)$)} 
\label{method:sensor}
The sensor-based failure is modeled to estimate probabilistic time-to-failure (i.e. probabilistic RUL) from time-series sensor measurements.

\textit{Mean RUL estimator}: Multivariate sensor time-series are segmented into sliding windows and provided as input to a 1D convolution neural network (1D-CNN).
The architecture consists of multi-layer 1D convolution networks, followed by a dense layer and then a single output neuron predicting a point estimate of RUL. All parts of the network utilize ReLU activation. The network uses mean square error as a loss function. The point estimates are used as $\hat{\mu}(S)$ of failure distribution from sensor measurements.

\textit{RUL uncertainty (std) estimator}: To obtain a predictive distribution rather than a point estimate, we model the conditional standard deviation as a function of the predicted RUL. A separate dense neural network with  two hidden layers and an output neuron, is trained to predict the absolute residual $|y - \hat{\mu}(\hat{y})|$. This network is trained with a Huber loss. The predictions from the dense network produces $\hat{\sigma}(S)$. 

The sensor-based time-to-failure distribution is modeled as \added{Gaussian, i.e.,} $P(F \mid S) = \mathcal{N}(\hat{\mu}(S), \hat{\sigma}(\hat{\mu}(S)))$ \added{and} truncated to non-negative remaining life. This defines the distribution $P(F \mid S)$ over the remaining time-to-failure $\tau$.


\subsection{Baseline failure distribution ($P(F)$)}

The baseline failure probability represents prior knowledge in the absence of both wear and sensor information. As mentioned in the Section \ref{Sec: ModelAssumptions}, the baseline distribution assumes constant hazard $\lambda_0 = 1 / MTBF$. This constant hazard yields an exponential distribution $f_0(t) = \lambda_0 \exp(-\lambda_0 t)$. This defines the distribution $P(F)$ over the remaining time-to-failure $\tau$. This is a memory-less model, it always predicts the same distribution irrespective of time, wear and senor information.



\subsection{\added{Fusion of components}}

To estimate the prediction of the combined model, individual prediction form different components $P(F \mid W)$, $P(F \mid S )$ and $P(F)$ are put together using the Equation (\ref{eq: assumption}). 

\added{In the derivation above, we use \(P(\cdot)\) as Bayesian notation for probability distributions. In the implementation, \(F\) is treated as a continuous remaining-life variable, and the three terms \(P(F \mid W)\), \(P(F \mid S)\), and \(P(F)\) are evaluated as probability density functions over future remaining life. Therefore, Equation~(1) first defines an unnormalized fused density,
\[
\tilde{P}(F \mid W,S)
=
\frac{P(F \mid W)P(F \mid S)}{P(F)}.
\]
The normalized fused density is then obtained by numerical normalization over the remaining-life grid:
\[
P(F_k \mid W,S)
=
\frac{\tilde{P}(F_k \mid W,S)}
{\sum_{j} \tilde{P}(F_j \mid W,S)\Delta F},
\]
where \(F_k \in \{0,1,\ldots,128\}\) cycles and \(\Delta F=1\) cycle in this work. }



\section{Experiments}
This section compiles the informataion on the dataset, experimental setup and the metrics used.

\subsection{Dataset and Experimental Setup}

\paragraph{Dataset:} We use the New Commercial Modular Aero-Propulsion System Simulation (N-CMAPSS) dataset \cite{arias2021aircraft}, which provides realistic simulated run-to-failure data of aircraft engines operating under real flight conditions. For each cycle a time-series is collected from take-off to landing. This time-series comprises of 14 sensor values, 14 virtual sensors and four operating conditions. This is collected until the unit fails.

N-CMAPSS consists of eight sub-datasets corresponding to different failure modes. In this study, we evaluate on: DS01, DS03 and DS04. Dataset DS02 was excluded because several time-series were shorter than the selected window length used for the 1D-CNN model. \added{Window length used in this work is 50 cycles.}

\paragraph{Experimental Setup:} The N-CMAPSS dataset provides pre-defined train and test splits. From the train set, we randomly selected two units for validation set for model development.

The wear-based component $P(F \mid W)$ is modeled using \textit{lifelines} package. Model parameters are estimated via maximum likelihood from training time-to-failure data,i.e., cycle at which a unit fails. Conditional RUL distributions are obtained by truncating and re-normalizing at the current cycle, as described in Section \ref{method:wear}




The sensor-based component $P(F\mid S)$ is implemented using 1D-CNN model. This model (i.e. $\hat{\mu}(S)$) uses three 1D-CNN layers (with 10, 10, and 1 filters) and followed by a dense layer with 100 neurons. The uncertainty model (i.e. $\hat{\sigma}(\hat{\mu}(S))$) has a single input neuron and output neuron with a hidden layer of 10 neurons. \removed{This standard deviation prediction network is trained on the validation set to as $\hat{\sigma}$ is trained as post processing step.}\added{The standard-deviation model is trained as a post-processing step using the validation set.} The resulting predictive distribution is Gaussian, as detailed in Section \ref{method:sensor}.

\subsection{Evaluation Metrics}
To comprehensively evaluate predictive performance, we consider both point-estimate and probabilistic metrics. The measures including Root Mean-Square Error (RMSE), Prediction Interval Coverage Probability (PICP), Normalized Mean Interval Prediction Width (NMIPW) and C-index. The measures are defined as follows:
\begin{itemize}
    \item RMSE measures the root mean squared error between predictions and ground truth values, evaluating the accuracy of point estimates (Equation \ref{eq: rmse}).
    
    \item PICP evaluates probabilistic predictions by measuring the proportion of samples for which the true value lies within the predicted interval (Equation \ref{eq: pcip}).
    
    \item NIPW measures the normalized average width of the prediction intervals across samples, indicating the sharpness of probabilistic predictions (Equation \ref{eq: NIPW}).
    
    \item C-index evaluates whether the ordering of predicted values is consistent with the ordering of the ground truth (Equation \ref{eq: cindex}).
\end{itemize}
\begin{align}
    RMSE(y, \hat{y}) = \added{ \sqrt{1/N \sum_{i} (y_i - \hat{y}_i)^2} }
    \label{eq: rmse}
\end{align}
\begin{align}
    PICP = \frac{1}{N} \sum_{i=1}^N \left\{\begin{array}{l}
                1 \text{ if } y_i \in [\added{L}_{\alpha}(\hat{\boldsymbol{p}}_i), \added{U}_{\alpha}(\hat{\boldsymbol{p}}_i)] \\
                0 \text{ if } y_i \notin [\added{L}_{\alpha}(\hat{\boldsymbol{p}}_i), \added{U}_{\alpha}(\hat{\boldsymbol{p}}_i)]
                \end{array}\right.\;
    \label{eq: pcip}
\end{align}
\begin{align}
    NIPW= \frac{1}{N(\max{\left\{y\right\}}-\min{\left\{y\right\}})} \sum_{i=1}^N (U_\alpha(\hat{\boldsymbol{p}}_i)-L_\alpha(\hat{\boldsymbol{p}}_i))
    \label{eq: NIPW}
\end{align}
\begin{equation}
\begin{aligned}
\text{C-index}(y, \hat{y})
&=
\added{
\frac{1}{n}
\sum_{i=1}^{N-1}
\sum_{j=i+1}^{N}
I(y_i < y_j)
} \\
&\quad \added{\times
\left[
I(\hat{y}_i < \hat{y}_j)
+
0.5\,I(\hat{y}_i = \hat{y}_j)
\right]}
\removed{
\frac{1}{N}
\sum_i \sum_j
I\bigl(\hat{y}_i > \hat{y}_j,\; y_i < y_j\bigr)
}
\end{aligned}
\label{eq: cindex}
\end{equation}

Here $y$ and $\hat{y}$ denotes the ground truth and estimated time to failure (i.e. RUL). $N$ denotes the number of samples in the dataset, $\hat{p}_i$ denotes the estimated probability distribution for the $i^{th}$ sample, \added{$I(.)$ denotes the indicator function, $n$ is the number of comparable pairs (calculated via $\sum_{i=1}^{N-1} \sum_{j=i+1}^{N} \mathbf{1}(y_i < y_j)$),} $U_\alpha(p)$ and $L_\alpha(p)$ provides the upper and lower bounds of the distribution $p$. In this work, we set \(\alpha = 95\%\). 

\section{Results}
\subsection{Empirical Evaluation}
\label{sec: evaluation}



We  evaluate the Weibull model ($P(F \mid W)$), 1D-CNN model ($P(F \mid S)$) and the proposed combined model ($P(F \mid W, S)$) across DS01, DS03, and DS04. Results averaged over 10 runs are reported in Table \ref{tab:results_summary}. \added{The final RUL point estimate used for RMSE is computed as the mean of the normalized fused density. Prediction intervals are computed from the normalized fused distribution by converting the desired confidence level \(\alpha\) to the corresponding standard-deviation multiplier and applying it to the fused mean and variance.}

\begin{table}[h!]
\caption{Performance comparison across datasets averaged over 10 runs. Methods include Weibull (W), 1D-CNN (1D), and the combined model (C). Values in parentheses indicate standard deviation, and arrows denote the direction of goodness for each metric.}
\label{tab:results_summary}
\resizebox{\columnwidth}{!}{
\begin{tabular}{llllll}
\toprule
 D & M  & RMSE$\downarrow$ & PICP$\uparrow$ & NIPW$\downarrow$ & C-index$\uparrow$ \\
\midrule
\multirow[b]{3}{*}{\rotatebox{90}{DS01}} 
 & W & 7.02 (0.00) & \textbf{1.00 (0.00)} & 0.34 (0.00) & \textbf{0.96 (0.00)} \\
 & 1D & 8.69 (1.06) & 0.80 (0.03) & 0.27 (0.05) & 0.90 (0.01) \\
 & C & \textbf{6.12 (0.23)} & 0.86 (0.04) & \textbf{0.21 (0.02)} & 0.94 (0.00) \\
\hline
\multirow[b]{3}{*}{\rotatebox{90}{DS03}} 
 & W & 9.24 (0.00) & \textbf{1.00 (0.00)} & 0.47 (0.00) & 0.88 (0.00) \\
 & 1D & 8.03 (0.55) & 0.89 (0.01) & 0.41 (0.02) & 0.90 (0.01) \\
 & C & \textbf{7.34 (0.30)} & 0.95 (0.01) & \textbf{0.30 (0.01)} & \textbf{0.91 (0.00)} \\
\hline
\multirow[b]{3}{*}{\rotatebox{90}{DS04}} 
 & W & 9.26 (0.00) & \textbf{1.00 (0.00)} & 0.46 (0.00) & \textbf{0.91 (0.00)} \\
 & 1D & 17.33 (1.59) & 0.85 (0.03) & 0.57 (0.08) & 0.78 (0.03) \\
 & C & \textbf{9.24 (0.39)} & 0.95 (0.03) & \textbf{0.35 (0.02)} & 0.90 (0.01) \\
\hline
\bottomrule
\end{tabular}
}
\end{table}

Across all datasets, the combined model achieves the lowest RMSE, indicating improved point prediction accuracy relative to both individual \removed{mdoels}\added{models}. This suggests that integrating aging-based and sensor-based information reduces overall prediction error.

The Weibull model attains the highest PICP values, reflecting conservative but well-calibrated prediction intervals. In particular, the reported PICP values of 1.00 (rounded to two decimals) reflect highly conservative prediction intervals induced by the parametric survival model. However, these intervals are comparatively wide, as reflected in the NIPW metric.

The combined model consistently produces narrower intervals than the Weibull model while maintaining competitive coverage, resulting in the best NIPW performance across datasets. This indicates improved sharpness without substantial loss of calibration.

In terms of ranking consistency (C-index), the Weibull model performs strongly due to its monotonic aging structure. The combined model achieves comparable C-index values, demonstrating that the fusion process preserves ordering performance while improving accuracy and interval sharpness.

Overall, the empirical results show that the proposed probabilistic fusion framework provides a favorable trade-off between accuracy, calibration, and sharpness compared to either model used independently.

\subsection{Visual Evaluation}

\begin{figure*}[t!]
    \centering
    \includegraphics[width=\textwidth]{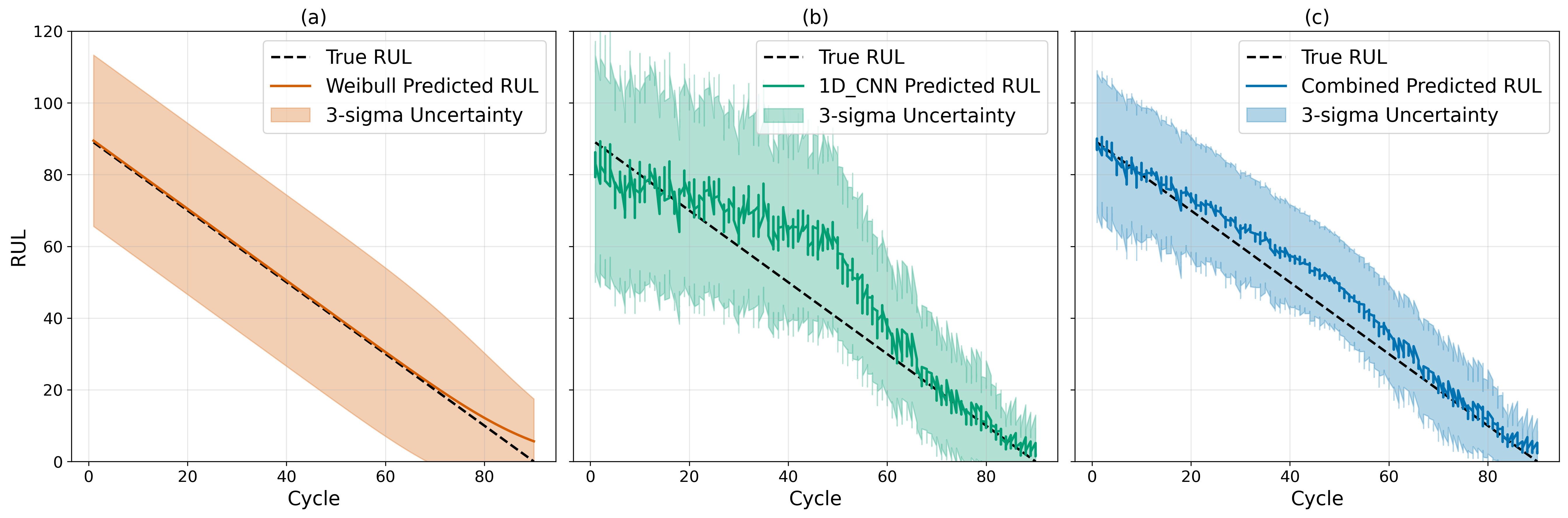}
    \caption{RUL predictions for \textit{Unit 7 from dataset DS01}. Panel (a) shows the Weibull survival model, (b) the sensor-based 1D-CNN model, and (c) the combined model. The dashed line denotes the true RUL, and shaded regions indicate the 3$\sigma$ uncertainty intervals.}
    \label{fig:plot_pred_ds01}
\end{figure*}
\begin{figure*}
    \centering
    \includegraphics[width=\textwidth]{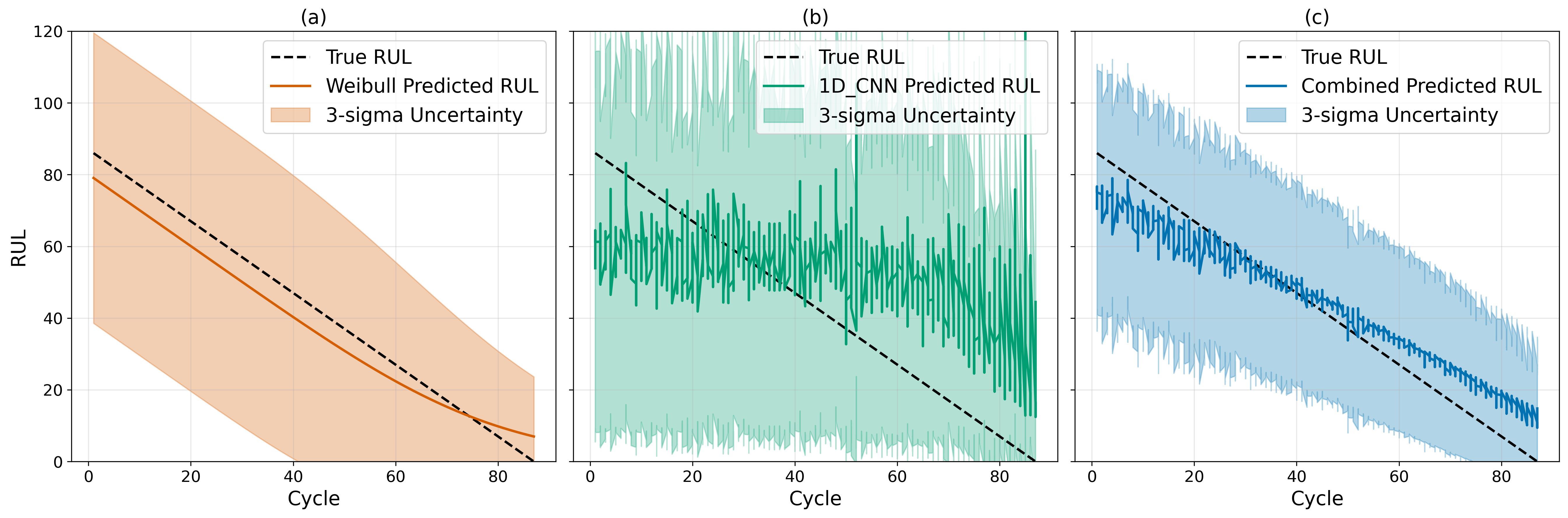}
    \caption{RUL predictions for \textit{Unit 7 from dataset DS04}. Panel (a) shows the Weibull survival model, (b) the sensor-based 1D-CNN model, and (c) the combined model. The dashed line denotes the true RUL, and shaded regions indicate the 3$\sigma$ uncertainty intervals.}
    \label{fig:plot_pred_ds04}
\end{figure*}


To complement the quantitative results, Figures \ref{fig:plot_pred_ds01} and \ref{fig:plot_pred_ds04} present representative RUL predictions for test unit $7$ from datasets DS01 and DS04, respectively. The DS01 example illustrates an ideal scenario in which both models produce accurate predictions, whereas the DS04 example highlights a more challenging case where the 1D-CNN model exhibits comparatively poorer predictive performance.

In Figure \ref{fig:plot_pred_ds01} prediction for test unit $7$ from dataset DS01, the Weibull model follows the true RUL trajectory closely for this unit. The 1D-CNN model initially exhibits large predictive variance when the system is far from failure, reflecting limited degradation information in the early stages of operation. As the unit approaches failure, the predictions become more accurate and the uncertainty decreases. The combined model integrates information from both components, producing predictions that remain stable in early cycles while gradually incorporating the improved sensor-based estimates later in the lifecycle.

In Figure \ref{fig:plot_pred_ds04} prediction for test unit $7$ from dataset DS04, the 1D-CNN model exhibits larger deviations and wider uncertainty bounds compared to the Weibull model. The Weibull model produces smoother predictions, with comparatively broad intervals.The combined model lies between the two estimates and shifts toward the Weibull prediction at large, while still utilizing the predictions from sensor-based model.

\subsection{Error Analysis of Combined Model}

\begin{figure}
    \centering
    \includegraphics[width=1\linewidth]{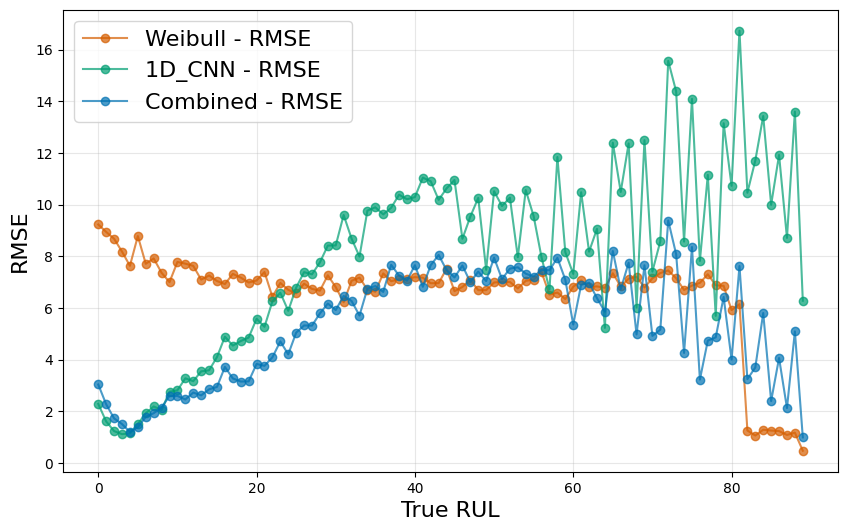}
    \caption{Prediction error as a function of true RUL for dataset DS01. The combined model follows the Weibull model in early life and shifts toward the sensor-based model as failure approaches.}
    \label{fig:error_plot_ds_01}
\end{figure}

To further analyze the behavior of our fusion framework, Figure \ref{fig:error_plot_ds_01} presents the average prediction error (RMSE) as a function of true RUL for dataset DS01. When the true RUL is large, the combined model closely follows the error profile of the Weibull model. As the unit approaches failure, the error of the 1D-CNN decreases, and the combined model correspondingly shifts toward the sensor-based behavior. This illustrates a transition behavior of the combined model.

 
\subsection{Ablation Study}

\begin{figure*}
    \centering
    \includegraphics[width=\textwidth]{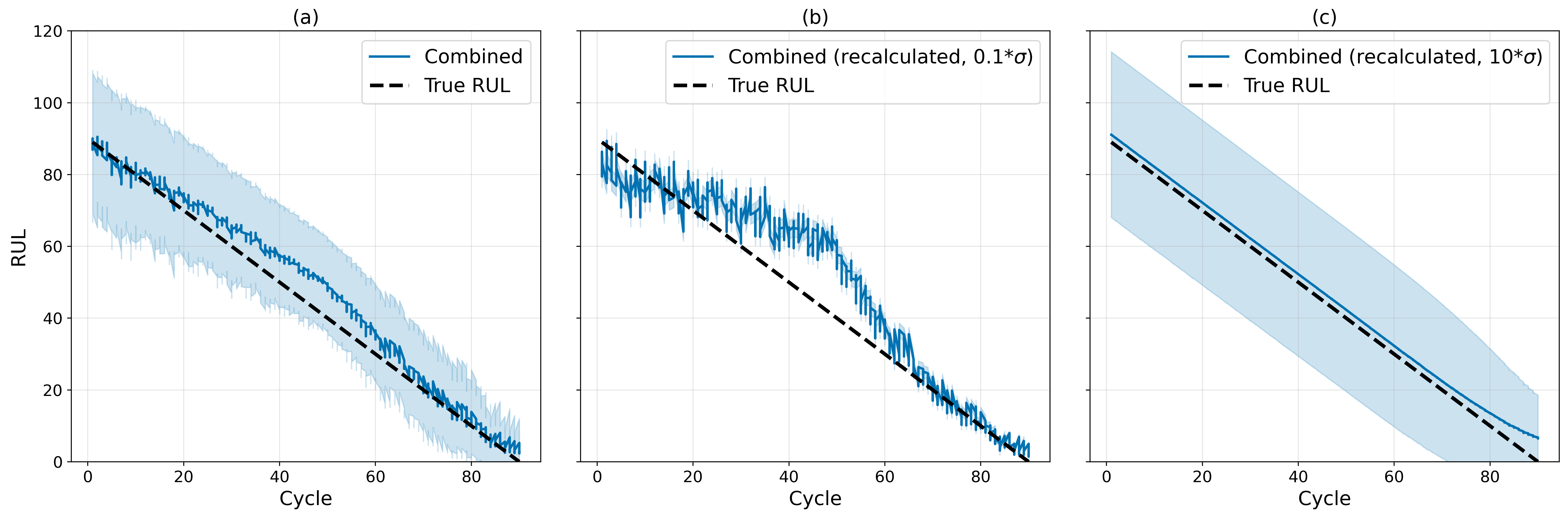}
    \caption{Ablation study showing the effect of predictive uncertainty of the sensor-based model on the combined model. Deflating the standard deviation of 1D-CNN model shifts the estimate toward the 1D-CNN prediction, while inflating it shifts the estimate toward the Weibull model. Original prediction can be seen in Figure \ref{fig:plot_pred_ds01}.}
    \label{fig:ablation}
\end{figure*}

To examine the influence of predictive uncertainty on the combined model, we synthetically scale the estimated standard deviation of the 1D-CNN while keeping the predicted mean fixed. The resulting predictions are shown in Figure \ref{fig:ablation}.

When the standard deviation is deflated, the combined estimate converges toward the 1D-CNN prediction. When the standard deviation is inflated, the combined estimate shifts toward the Weibull model. The prediction \removed{form}\added{from} individual models can be seen in Figure \ref{fig:plot_pred_ds01}. This experiment demonstrates that the combined model's outcome varies systematically with the relative uncertainty of the sensor-based component. Ablation Study solidifies the transition behaviors observed in the error analysis.
 


\section{Discussion}

The empirical results demonstrate that integrating wear-based survival modeling with sensor-based RUL estimation yields consistent improvements across datasets. The combined model achieves lower RMSE while maintaining competitive c-index and producing narrower prediction intervals than either component individually. These results indicate that the fusion framework provides a favorable balance between accuracy, calibration, and interval width.

The visual analysis further clarifies this behavior. When the sensor-based model exhibits comparatively large uncertainty, its influence is limited, and the combined prediction remains closer to the Weibull estimate. At the same time, sensor-derived information is still incorporated into the final estimate. This adaptive behavior suggests that the fusion mechanism performs uncertainty-aware decision making rather than fixed model averaging.

The error analysis provides additional insight into this adaptation. In early stages of operation, when the true RUL is large, sensor signals contain limited degradation information. In this regime, the wear-based model offers stable and consistent predictions, and the combined model closely follows its behavior. As failure approaches and degradation becomes more observable in sensor measurements, the sensor-based model improves in accuracy. The combined model correspondingly shifts toward the sensor-driven estimate. This time-dependent transition emerges naturally from the fusion rule.

The ablation study reinforces this interpretation. By synthetically inflating or deflating the estimated variance of the sensor-based model, the dominance of each component changes systematically. When the sensor-based model is confident (low variance), it dominates the combined estimate. When its uncertainty increases, the wear-based prior exerts greater influence. This confirms that the switching behavior is governed by relative predictive uncertainty and aligns directly with the probabilistic formulation in Equation (\ref{eq: assumption}). The observed behavior therefore arises from principled probabilistic integration rather than heuristic blending.

The effectiveness of this framework depends critically on well-calibrated uncertainty estimates. If predictive variances are substantially mis-calibrated, the fusion mechanism may overweight unreliable predictions or underweight reliable predictions. Consequently, uncertainty calibration plays an important role in practical deployment.


This study focuses on a specific parametric survival model (i.e. the Weibull) and a specific neural architecture (i.e. 1D-CNN) for sensor-based modeling to evaluate our primary contribution on the probabilistic fusion mechanism. In principle, our framework is model-agnostic and can accommodate alternative survival formulations, such as Cox-based, stochastic degradation models, or more advanced deep learning architectures. Evaluating the approach with a range of individual models would provide further insight into its generality and the extent of performance gains in relation to individual- model performance.


The experimental evaluation is conducted on the N-CMAPSS dataset\removed{, which is based on simulated component degradation. While this allows controlled analysis, real-world data often exhibit additional complexities and noise characteristics that may challenge model performance.}\removed{verion2: The experimental evaluation is conducted on N-CMAPSS, a simulated dataset that enables controlled analysis of the proposed fusion mechanism but may also limit the extent to which the empirical findings generalize to real-world fleets. Because N-CMAPSS is generated from an underlying simulation-based degradation process, its degradation trajectories may be more structured and statistically regular than those observed in operational data. Real-world assets often involve unobserved maintenance actions, heterogeneous usage histories, changing operating regimes, sensor drift, missing data, and multiple interacting degradation mechanisms.}\added{, a simulated dataset, that enables controlled analysis of the proposed fusion mechanism but may not fully reflect the complexity of real-world fleets. Its simulation-based degradation process may produce more structured and statistically regular trajectories. In contrast, in real world operational data additional variability is expected to be introduced through maintenance actions, usage histories, operating regimes, sensor drift, missing data, and interacting degradation mechanisms.} Assessing the proposed framework on real industrial datasets remains an important direction for future work. 


\removed{Finally, the framework relies on a conditional independence assumption between wear and sensor information given failure.} \removed{\added{Finally, the framework relies on an additional conditional independence approximation between wear and sensor information given failure, motivated by the use of failure-time distributions as observable proxies for the latent health state. This approximation is reasonable as a working assumption in the controlled N-CMAPSS setting, where degradation, health evolution, sensor responses, and failure are generated by an underlying simulation mechanism. However,} in practice, this assumption may be violated, particularly}
\added{The framework relies on two assumptions, introduced in section 3: First, that there is a deterministic relation between H and F, and second that W and S are independent given H. Care must be taken when applying the proposed method, since both assumptions may be violated in certain situations. First, the assumption of equivalence between H and F relies on that the expression for equivalent operating hours used in modeling $P(F)$ corresponds to the actual physical degradation mechanism in the component. 
For example, if the actual degradation accumulates as a function of load, but remaining life is modeled in calendar time (and the future load is not known in advance), then 
the relation between H and F is not deterministic any more (only correlated) and the combined model will be an approximation. Second, the assumption that W and S are independent given H may also be violated, if for example information about the wear leaks into the sensor based prediction. This may happen} when wear-related variables are implicitly encoded in sensor measurements (e.g., trip meters, fuel consumption, or time since last maintenance). Although such signals may benefit standalone sensor-based models, our results suggest that separating these sources and combining them through a principled survival model may be more effective. Future work should therefore investigate methods to test this assumption and develop strategies to account for potential dependencies when it does not hold.

\section{Conclusion}
This work proposed a probabilistic framework for integrating wear-based survival modeling with sensor-based RUL estimation under explicit structural assumptions. By modeling wear-driven risk and sensor-driven degradation as complementary sources of information. Later, combining the predictions at the level of failure probability distributions, the framework enables principled uncertainty-aware fusion rather than heuristic model averaging.

Empirical evaluation on N-CMAPSS datasets demonstrates that the combined model achieves improved prediction accuracy while maintaining competitive c-index performance and producing narrower prediction intervals. The error analysis further reveals a transition behavior: wear-based modeling provides stability in early life, whereas sensor-based estimation dominates as degradation becomes observable. This switching behavior emerges naturally from the probabilistic formulation and is governed by relative predictive uncertainty.


The proposed framework provides a structured approach to bridging wear-based and sensor-based prognostics, while also connecting industrial maintenance practices with predictive maintenance research. In industrial settings, long-term service plan are required to be stable, for which survival-based approaches are well suited. In contrast, much of the research literature emphasizes data-driven deep learning models for RUL estimation.\removed{By enabling}\added{ The proposed framework enables} the integration of advanced data-driven models with existing survival-based approaches, rather than replacing them. \added{Thus} the proposed framework supports a gradual and practical transition toward more data-driven and AI-enabled maintenance strategies.

\section*{Acknowledgment}
This work was supported by funding from Vinnova through the project \textit{Future of AI-Based Maintenance} (project number 2023-01917). We thank Jan Ekman, Pontus Slottner, and Sara Gestrelius for their valuable discussions.

The authors used ChatGPT-5.2 to refine the manuscript's language, improve clarity, and ensure grammatical correctness. The final content was reviewed and edited by the authors, who take full responsibility for the integrity of the scientific findings.

\bibliographystyle{apacite}
\begin{singlespace}
\bibliography{ijphm}

@article{arias2021aircraft,
  title={Aircraft engine run-to-failure dataset under real flight conditions for prognostics and diagnostics},
  author={Arias Chao, Manuel and Kulkarni, Chetan and Goebel, Kai and Fink, Olga},
  journal={Data},
  volume={6},
  number={1},
  pages={5},
  year={2021},
  publisher={MDPI}
}

@inproceedings{rahat2024survloss,
  title={Survloss: A new survival loss function for neural networks to process censored data},
  author={Rahat, Mahmoud and Kharazian, Zahra},
  booktitle={PHM Society European Conference},
  volume={8},
  number={1},
  pages={7--7},
  year={2024}
}

@article{katzman2018deepsurv,
  title={DeepSurv: personalized treatment recommender system using a Cox proportional hazards deep neural network},
  author={Katzman, Jared L and Shaham, Uri and Cloninger, Alexander and Bates, Jonathan and Jiang, Tingting and Kluger, Yuval},
  journal={BMC medical research methodology},
  volume={18},
  number={1},
  pages={24},
  year={2018},
  publisher={Springer}
}

@article{ishwaran2008random,
  title={Random survival forests},
  author={Ishwaran, Hemant and Kogalur, Udaya B and Blackstone, Eugene H and Lauer, Michael S},
  year={2008}
}

@article{kaplan1958nonparametric,
  title={Nonparametric estimation from incomplete observations},
  author={Kaplan, Edward L and Meier, Paul},
  journal={Journal of the American statistical association},
  volume={53},
  number={282},
  pages={457--481},
  year={1958},
  publisher={Taylor \& Francis}
}

@article{cox1972regression,
  title={Regression models and life-tables},
  author={Cox, David R},
  journal={Journal of the royal statistical society: Series B (methodological)},
  volume={34},
  number={2},
  pages={187--202},
  year={1972},
  publisher={Wiley Online Library}
}

@article{fink2020potential,
  title={Potential, challenges and future directions for deep learning in prognostics and health management applications},
  author={Fink, Olga and Wang, Qin and Svensen, Markus and Dersin, Pierre and Lee, Wan-Jui and Ducoffe, Melanie},
  journal={Engineering Applications of Artificial Intelligence},
  volume={92},
  pages={103678},
  year={2020},
  publisher={Elsevier}
}

@inproceedings{zheng2017long,
  title={Long short-term memory network for remaining useful life estimation},
  author={Zheng, Shuai and Ristovski, Kosta and Farahat, Ahmed and Gupta, Chetan},
  booktitle={2017 IEEE international conference on prognostics and health management (ICPHM)},
  pages={88--95},
  year={2017},
  organization={IEEE}
}

@article{nemani2022ensembles,
  title={Ensembles of probabilistic LSTM predictors and correctors for bearing prognostics using industrial standards},
  author={Nemani, Venkat P and Lu, Hao and Thelen, Adam and Hu, Chao and Zimmerman, Andrew T},
  journal={Neurocomputing},
  volume={491},
  pages={575--596},
  year={2022},
  publisher={Elsevier}
}

@article{zhang2023degradation,
  title={Degradation modeling and RUL prediction with Wiener process considering measurable and unobservable external impacts},
  author={Zhang, Shuyi and Zhai, Qingqing and Li, Yaqiu},
  journal={Reliability Engineering \& System Safety},
  volume={231},
  pages={109021},
  year={2023},
  publisher={Elsevier}
}

@article{zezhou2024stochastic,
  title={Stochastic degradation modeling and remaining useful lifetime prediction based on long short-term memory network},
  author={Zezhou, WANG and Jian, Hou and Jiantai, Zhu and Liyuan, Wang and Zhongyi, Cai},
  journal={Measurement},
  volume={234},
  pages={114803},
  year={2024},
  publisher={Elsevier}
}

@inproceedings{bechhoefer2008use,
  title={Use of Paris law for prediction of component remaining life},
  author={Bechhoefer, Eric and Bernhard, Andreaus and He, David},
  booktitle={2008 IEEE Aerospace Conference},
  pages={1--9},
  year={2008},
  organization={IEEE}
}

@article{murtaza2024paradigm,
  title={Paradigm shift for predictive maintenance and condition monitoring from Industry 4.0 to Industry 5.0: A systematic review, challenges and case study},
  author={Murtaza, Aitzaz Ahmed and Saher, Amina and Zafar, Muhammad Hamza and Moosavi, Syed Kumayl Raza and Aftab, Muhammad Faisal and Sanfilippo, Filippo},
  journal={Results in Engineering},
  volume={24},
  pages={102935},
  year={2024},
  publisher={Elsevier}
}

@article{nieves2022semi,
  title={A semi-supervised approach with monotonic constraints for improved remaining useful life estimation},
  author={Nieves Avendano, Diego and Vandermoortele, Nathan and Soete, Colin and Moens, Pieter and Ompusunggu, Agusmian Partogi and Deschrijver, Dirk and Van Hoecke, Sofie},
  journal={Sensors},
  volume={22},
  number={4},
  pages={1590},
  year={2022},
  publisher={MDPI}
}

@article{cao2024hybrid,
  title={A hybrid data-and model-driven learning framework for remaining useful life prognostics},
  author={Cao, Hongjie and Xiao, Wei and Sun, Jian and Gan, Ming-Gang and Wang, Gang},
  journal={Engineering Applications of Artificial Intelligence},
  volume={135},
  pages={108557},
  year={2024},
  publisher={Elsevier}
}

@book{klein2003survival,
  title={Survival analysis: techniques for censored and truncated data},
  author={Klein, John P and Moeschberger, Melvin L},
  volume={1230},
  year={2003},
  publisher={Springer}
}

@article{deng2015degradation,
  title={Degradation modeling based on a time-dependent ornstein-uhlenbeck process and residual useful lifetime estimation},
  author={Deng, Yingjun and Barros, Anne and Grall, Antoine},
  journal={IEEE Transactions on Reliability},
  volume={65},
  number={1},
  pages={126--140},
  year={2015},
  publisher={IEEE}
}

@article{cygu2023comparing,
  title={Comparing machine learning approaches to incorporate time-varying covariates in predicting cancer survival time},
  author={Cygu, Steve and Seow, Hsien and Dushoff, Jonathan and Bolker, Benjamin M},
  journal={Scientific reports},
  volume={13},
  number={1},
  pages={1370},
  year={2023},
  publisher={Nature Publishing Group UK London}
}

@article{yang2022prognostic,
  title={Prognostic modeling of predictive maintenance with survival analysis for mobile work equipment},
  author={Yang, Zhen and Kanniainen, Juho and Krogerus, Tomi and Emmert-Streib, Frank},
  journal={Scientific reports},
  volume={12},
  number={1},
  pages={8529},
  year={2022},
  publisher={Nature Publishing Group UK London}
}

@inproceedings{rahat2023bridging,
  title={Bridging the gap: A comparative analysis of regressive remaining useful life prediction and survival analysis methods for predictive maintenance},
  author={Rahat, Mahmoud and Kharazian, Zahra and Mashhadi, Peyman Sheikholharam and R{\"o}gnvaldsson, Thorsteinn and Choudhury, Shamik},
  booktitle={Phm society asia-pacific conference},
  volume={4},
  number={1},
  year={2023}
}

@article{ Zhang_2025jaciii,
title={Remaining Useful Life Prediction for Tools Based on Monitoring Data and Stochastic Degradation Model},
author={Baokang Zhang and Ning Li and Jiahui Huang and Takahiro Arakawa and Kentaro Ishii and Ryuichi Yashima},
journal={Journal of Advanced Computational Intelligence and Intelligent Informatics},
volume={29},
number={3},
pages={668-676},
year={2025},
doi={10.20965/jaciii.2025.p0668}
	}

@article{liao2016hybrid,
  title={A hybrid framework combining data-driven and model-based methods for system remaining useful life prediction},
  author={Liao, Linxia and K{\"o}ttig, Felix},
  journal={Applied Soft Computing},
  volume={44},
  pages={191--199},
  year={2016},
  publisher={Elsevier}
}

@article{chao2022fusing,
  title={Fusing physics-based and deep learning models for prognostics},
  author={Chao, Manuel Arias and Kulkarni, Chetan and Goebel, Kai and Fink, Olga},
  journal={Reliability Engineering \& System Safety},
  volume={217},
  pages={107961},
  year={2022},
  publisher={Elsevier}
}
\end{singlespace}

\section*{Appendix}
\subsection*{Motivation on the standard deviation Model}
\begin{figure*}[ht]
    \centering
    \includegraphics[width=0.4\textwidth, trim= 0 0 0 0.8cm, clip]{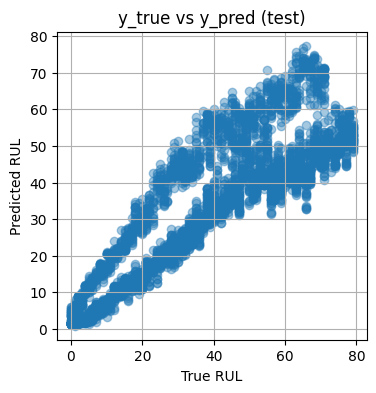}
    \caption{Scatter plot of true versus predicted RUL for the 1D-CNN model. The increasing spread at larger RUL values motivates modeling heteroskedastic prediction uncertainty.}
    \label{fig:std_model_motivation}
\end{figure*}

Scatter plot in Figure \ref{fig:std_model_motivation} show that the true RUL vs predicted RUL does not perfectly align on the diagonal line and additionally has a spread around the diagonal line. This calls for modeling the heteroskedasticity uncertainty of the model. We addressed this by modeling this uncertainty term as $\hat{\sigma}(\hat{\mu}(S))$.
\subsection*{Validating the results across individual Units}

\begin{figure*}
    \centering
    \includegraphics[width=0.95\textwidth]{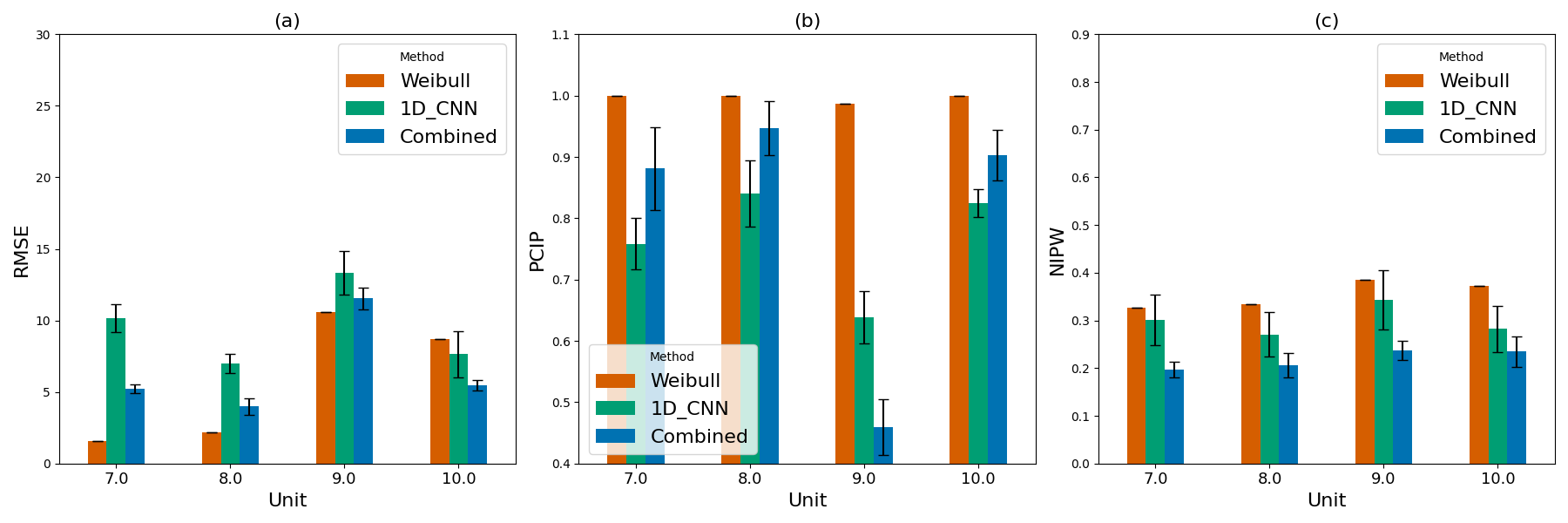}
    \caption{Per-unit evaluation of predictive performance, verifying that overall trends are consistent across individual units for the dataset DS01.}
    \label{fig:result_per_unit}
\end{figure*}

To assess the consistency of the results, we evaluate performance across individual units. Figure \ref{fig:result_per_unit} presents the metrics computed per unit, demonstrating that the overall trends observed in Section \ref{sec: evaluation} are preserved at the unit level. In particular, the combined model consistently achieves the best performance in terms of NIPW, while maintaining comparable RMSE and PCIP\added{, with a exception on unit 9 with lower PCIP}. Similar behavior is observed across all evaluated datasets. 

\added{For Unit 9, the combined model shows a noticeable reduction in PICP compared with the other units. This behavior is mainly attributed to local miscalibration of the predictive uncertainty for this unit. In particular, increasing the confidence level used for PICP from \(\alpha=0.95\) to \(\alpha=0.99\) raises the PICP for Unit 9 to 0.81. This indicates that the prediction errors are not entirely inconsistent with the estimated distribution, but that the 95\% predictive interval is too narrow for this specific unit. Therefore, the degraded PICP for Unit 9 reflects local uncertainty miscalibration.}

\subsection*{Additional Information about the data}

\added{Our experiments use three N-CMAPSS subsets: DS01 (4 train, 2 validation, 4 test units), DS03 (7 train, 2 validation, 6 test units), and DS04 (4 train, 2 validation, 4 test units). For each cycle, the time-series data are smoothed using a 20-step tumbling-window average and segmented using a sliding window of length 50. This produces DS01 (14,359 training, 7,728 validation, 12,156 test samples), DS03 (20,567 training, 4,369 validation, 19,312 test samples), and DS04 (19,386 training, 10,241 validation, 16,489 test samples).}






\end{document}